\documentclass{article}

     \usepackage[preprint]{neurips_2020}

\usepackage[utf8]{inputenc} 
\usepackage[T1]{fontenc}    
\usepackage{hyperref}       
\usepackage{url}            
\usepackage{booktabs}       
\usepackage{amsfonts}       
\usepackage{nicefrac}       
\usepackage{microtype}      

\usepackage{amsthm}
\usepackage{amssymb}
\usepackage{color}
\usepackage{multirow}

\usepackage{booktabs}
\usepackage{graphicx}
\usepackage{xcolor}
\usepackage{wrapfig}
\usepackage[normalem]{ulem}

\usepackage{amsmath,amsfonts,bm}

\def\eqref#1{equation~\ref{#1}}

\def\floor#1{\lfloor #1 \rfloor}
\def\1{\bm{1}}

\DeclareMathAlphabet{\mathsfit}{\encodingdefault}{\sfdefault}{m}{sl}
\SetMathAlphabet{\mathsfit}{bold}{\encodingdefault}{\sfdefault}{bx}{n}

\title{Perception-Prediction-Reaction Agents for Deep Reinforcement Learning}

\author{
  Adam Stooke\thanks{Work done while at DeepMind, London, UK.} \\
  Department of Computer Science\\
  University of Berkeley\\
  \texttt{adam.stooke@berkeley.edu} \\
  \And
  Valentin Dalibard \\
  DeepMind\\
  \And
  Siddhant M.  Jayakumar\\
  DeepMind\\
  \And
  Wojciech M. Czarnecki\\
  DeepMind\\
  \And
  Max Jaderberg\\
  DeepMind\\
}

\begin{document}

\maketitle

\begin{abstract}
We introduce a new recurrent agent architecture and associated auxiliary losses which improve reinforcement learning in partially observable tasks requiring long-term memory.  We employ a temporal hierarchy, using a slow-ticking recurrent core to allow information to flow more easily over long time spans, and three fast-ticking recurrent cores with connections designed to create an information asymmetry.  The \emph{reaction} core incorporates new observations with input from the slow core to produce the agent's policy; the \emph{perception} core accesses only short-term observations and informs the slow core; lastly, the \emph{prediction} core accesses only long-term memory.  An auxiliary loss regularizes policies drawn from all three cores against each other, enacting the prior that the policy should be expressible from either recent or long-term memory.  We present the resulting \emph{Perception-Prediction-Reaction} (PPR) agent and demonstrate its improved performance over a strong LSTM-agent baseline in DMLab-30, particularly in tasks requiring long-term memory.  We further show significant improvements in Capture the Flag, an environment requiring agents to acquire a complicated mixture of skills over long time scales.  In a series of ablation experiments, we probe the importance of each component of the PPR agent, establishing that the entire, novel combination is necessary for this intriguing result.

\end{abstract}

\section{Introduction}
In the reinforcement learning (RL) problem, an agent is trained to solve an environment cast as a Markov decision process (MDP), specified as a tuple of states, actions, transition probabilities, and rewards:  $(\mathcal{S}, \mathcal{A}, P, r)$.  The agent must learn which states and actions lead to the best rewards without prior knowledge of $P$.  In many interesting RL problems, however, the agent receives an observation, $x_t=o(s_t) \in \mathcal{X}$, which does not completely specify the state of the MDP at that time step, resulting in \emph{partial observability}.  Therefore, for partially observable Markov decision processes (POMDPs)~\citep{astrom1965optimal,Kaelbling1998}, a focus of agent design is how to integrate the sequence of historical observations $(x_0, x_1, \ldots, x_t)$ to best approximate the state $s_t$ and produce a policy $\pi_t$ to maximise future rewards.  In deep RL, recurrent neural networks (RNNs) allow integrating observations over time with constant computational complexity~\citep{mnih2016asynchronous}.  Agents based on traditional recurrent networks, \emph{e.g.} LSTMs~\citep{hochreiter1997long}, are widely effective, but they sometimes struggle to learn in more complex environments involving long-term memory retention.

In this paper, we introduce a recurrent agent architecture, and associated auxiliary losses~\citep{jaderberg2016reinforcement}, which aim to improve deep RL in partially observable environments, particularly those requiring long-term memory. 
Specifically, we introduce a slowly ticking recurrent core to augment the standard fast ticking agent core, to allow a pathway for long-term memory storage and ease the backwards flow of gradients over long time spans. In addition, we construct two auxiliary policies, the first of which is required to use only current observations without long-term memory (\emph{perception}), and the second which must only use the long-term memory without current observations (\emph{prediction}). These auxiliary policies are trained jointly with the full-information policy (\emph{reaction}), with all three policies regularizing each other and shaping the representation of the slow-ticking recurrent core.

We evaluate this agent, dubbed the \emph{Perception-Prediction-Reaction} agent (PPR) on a suite of experiments on 3D, partially observable environments~\citep{DMLab}, and show consistent improvement compared to strong baselines, in particular on tasks requiring long-term memory. Ablation studies highlight the efficacy of each of the structural priors introduced in this paper. Finally, we apply this agent to the challenging DMLab-30 domain (one agent which must learn across 30 different POMDPs simultaneously) and Capture the Flag~\citep{jaderberg2018human}, and show that even in these highly varied RL domains, the PPR agent can improve performance.  It is striking that our simple regularising losses can have such a strong effect on learning dynamics.  Without intending to replicate any biological system, we drew loose inspiration for our architecture from concepts in hierarchical sensorimotor control; we refer the interested reader to related articles such as~\citep{loeb1999hierarchical,ting2007dimensional,ting2015neuromechanical,merel2019hierarchical}.

\section{The Perception-Prediction-Reaction Agent}

\begin{figure}[t]
\centering
\includegraphics[width=0.8\textwidth]{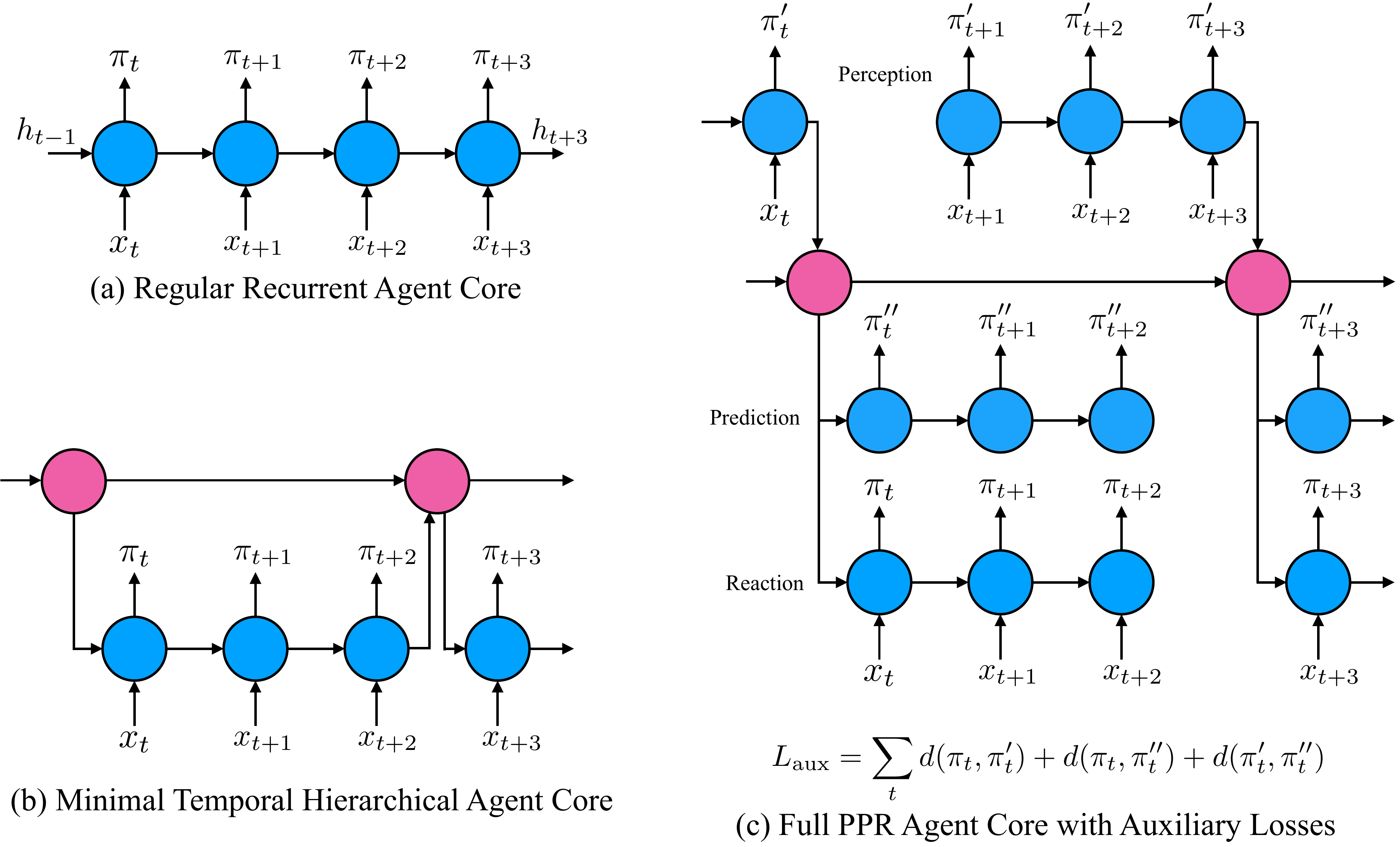}
\caption{\small (a) A regular recurrent agent core~\citep{mnih2016asynchronous} which can integrate historical experience of observations $x_t$ using an RNN to produce a policy $\pi_t$. (b) A minimal temporal hierarchical agent core, featuring fast- and slow-ticking recurrences~\citep{jaderberg2018human}. The slow-ticking core skips large portions of time, facilitating BPTT. (c) The PPR agent recurrent structure introduced in this paper, featuring a slow-ticking core and three fast-ticking cores. The \emph{perception} and \emph{prediction} fast-ticking branches have different information hidden relative to the \emph{reaction} fast-ticking core, which has full information and produces the behaviour policy.  All fast cores can share the same NN weights. An auxiliary loss $L_{\text{aux}}$ encourages the fast-ticking branches to predict the same policy with different information, where $d$ is the symmetrized Kullback-Leibler Divergence.}
\label{fig:recurrent}
\end{figure}

This section introduces the structural and objective priors which constitute the PPR agent.  We start with background on recurrent neural network-based agents for reinforcement learning, followed by discussion of a minimal hierarchical agent as an intermediate concept.

\subsection{Reinforcement Learning and Recurrent Agents}

In an MDP, the goal of the RL agent is to find a policy over actions, $\pi(a_t|s_t)$, conditioned on the state, that maximizes the expected discounted sum of future rewards, $\mathbb{E}_{s_t,a_t\sim\pi,P}[\sum_{t=0}^{\infty}{\gamma^tr_t}]$.  The objective remains the same under partial observability (POMDPs) however the agent does not have access to the full state, instead receiving incomplete information gleaned from observations, $x_t = o(s_t)\in \mathcal{X}$.  

In POMDPs, recurrent agents can improve their internal understanding of the current state by carrying information from past observations, $(x_0, x_1, \ldots, x_{t-1})$, in an internal state, $h_{t-1}$, to complement the current observation, $x_t$.  The agent updates its internal state by $h_t=f(x_t,h_{t-1})$, and the remaining network layers receive conditioning as $g(a_t| h_t)$ to produce the policy, see Figure~\ref{fig:recurrent}~(a).  Training by backpropagation through time (BPTT) ~\citep{Webros1990BACKPROPAGATIONTT,Rumelhart1988LearningIR} allows rewards to influence the processing of observations and internal state over earlier time steps.  Sophisticated recurrent functions, $f$, can extend the agent's ability to handle longer (and hence more difficult) sequences.  LSTM-based agents have succeeded in a range of partially observable environments, including ones with rich visual observations ~\citep{mnih2016asynchronous, IMPALA}, but many such tasks remain difficult to master or are learned slowly with the traditional architecture.

\subsection{Minimal Temporally Hierarchical Agent}
\label{sec:HVA}

Temporal hierarchy promises to further improve the processing of long sequences by dividing responsibilities for short- and long-term memory over different recurrent cores, simplifying the roles of each.  See Related Work for numerous examples of architectures with different hidden neurons operating at different time scales.  As a special case of this concept applied to RL, we consider employing an additional recurrent unit operating at a rate slower than the MDP.  This unit reduces the number of intermediate computations between distant time steps and allows error gradients to skip backwards through long segments of time.  

An example of such a hierarchical agent is shown in Figure~\ref{fig:recurrent}~(b): the slow core advances every $\tau$ time steps (depicted is $\tau=3$); during the interim it provides a fixed output to modulate the fast core; the fast core provides summary information to the slow core.  As depicted, the recurrence equations could take the following form:
\begin{equation}
\label{eq:minimal}
h_t^S = \begin{cases}f_S(h_{t-1}^F, h_{t-1}^S) & \text{if } t\bmod{\tau}=0 \\ h_{t-1}^S & \text{otherwise}\end{cases} 
\quad 
h_t^F = \begin{cases}f_F(x_t, h_t^S, \emptyset) & \text{if } t\bmod{\tau}=0 \\ f_F(x_t, h_t^S, h_{t-1}^F) & \text{otherwise}\end{cases}
\end{equation}
where the superscripts $S$ and $F$ denote slow and fast cores, respectively, $h^S_t, h_t^F$ are the recurrent states, $x_t$ is the observation, and $\emptyset$ denotes a vector of zeros (\emph{i.e.} the initial recurrent state).  The policy could generically depend on the recurrent states, $\pi_t=g(h_t^F, h_t^S)$ (in our case $g$ is an MLP).  The internal state of the fast core is periodically reset to $\emptyset$ so as to divide memory responsibilities by time-scale; all information originating prior to $\tau\floor{t/\tau}$ must have routed through the slow core.  

This minimal hierarchical agent does not on its own guarantee efficient training of long-term memory in a way that improves overall learning relative to the flat agent, see ablations in Figure~\ref{fig:fig3}~(b). Indeed, previous examples of temporally hierarchical agents~\citep{VezhnevetsFun,jaderberg2018human} introduce auxiliary objectives to best make use of similar hierarchical structures.  

\subsection{The PPR Agent}
\label{sec:ppr_description}


Our contribution is to construct a new hierarchical structure and associated auxiliary losses which enhance recurrent learning.  It relies on architectural elements designed to create an information asymmetry (as in \cite{galashov2018information}) which permits leveraging certain policy priors to simultaneously (a) shape the representation of the slow-ticking core, (b) maximize information extracted from observations, and (c) balance the importance of both in the policy.

The PPR agent is depicted in Figure~\ref{fig:recurrent}~(c), and we build its description starting from the minimal hierarchical agent.  First, we eliminate the possibility of a trivial feed-through connection from fast-slow-fast. Rather than attempt a partial information bottleneck, we prevent the fast core (\emph{reaction}) which receives input from the slow core, from passing any output back to the slow core.  We introduce another fast-ticking core (\emph{perception}) which feeds its output into the slow core but does not take input from it.  Resetting the fast internal states at the interval $\tau$ forms branches in the graph. The \emph{reaction} branch produces the agent's behavior policy by integrating new observations together with the slow core's output.  The slow core assumes a central role in representing information originating prior to $\tau\floor{t/\tau}$, as it receives periodic, short-term summaries from the \emph{perception} branch, which also integrates observations.  This forms a Perception-Reaction Agent \emph{without auxiliary losses}, a baseline in our ablation experiments.

The final architectural element of the PPR agent is an additional fast recurrent core (\emph{prediction}).  It branches simultaneously to \emph{reaction} and receives the slow core's output and possibly partial information $p_t$.
This creates an information asymmetry against the \emph{perception} branch, which lacks long-term memory, and the fully-informed \emph{reaction} branch.  We can leverage this asymmetry to enhance recurrent learning.
We do so by drawing auxiliary policies, $\pi'$ and $\pi''$ from the \emph{perception} and \emph{prediction} branches, respectively, to form the auxiliary loss: 
\begin{equation}
    \label{eq:aux}
    \mathcal{L}_{aux}=\sum_t{d(\pi_t,\pi_t') +  d(\pi_t,\pi_t'') + d(\pi_t',\pi_t''})
\end{equation}
where $d$ is a statistical distance -- we use the symmetrized Kullback-Leibler Divergence.  All three branches are regularized against each other; $\mathcal{L}_{aux}$ encourages their policies to agree as much as possible despite their differences in access to information.  Rather than apply a loss directly on the recurrent state, which may assume somewhat arbitrary values, the policy distribution space offers grounding in the environment. 

The recurrence equations and policy structure of the PPR agent are summarized in Table~\ref{tab:recur_equations}. Loosely speaking, \emph{reaction} is a short-term sensory-motor loop, \emph{perception} a sensory loop, \emph{prediction} a motor loop, and the slow core a long-term memory loop, all of which are decoupled in forward operation. The auxiliary divergence losses can be seen as imposing two priors on the fully informed \emph{reaction} branch -- that the policy should be expressible from only recent observations (\emph{perception}) and from only long-term memory (\emph{prediction}).

\begin{table}[t]
\caption{\small Recurrence and policy equations of the PPR agent.}
\label{tab:recur_equations}
\small
\begin{center}
\renewcommand{\arraystretch}{1.4}
\begin{tabular}{@{}lrlll@{}}
\toprule
\multicolumn{1}{c}{\bf Core}  &\multicolumn{3}{c}{\bf Recurrence Equation}  &\multicolumn{1}{c}{\bf Policy} \\
\hline
 & & \underline{if $t\bmod{\tau}=0$}: & \underline{otherwise}: & \\
Slow & $h_t^S=$ & $\{f_S(h_t', h_{t-1}^S;\ \theta^S),$ & $  h_{t-1}^S\}$ & \\
Reaction & $h_t=$ & $\{f(x_t, h_t^S, \emptyset;\ \theta),$ & $f(x_t, h_t^S, h_{t-1};\ \theta)\}$ & $\pi_t = g(h_t;\ \phi)$ \\

Prediction & $h_t''=$ & $\{f(p_t, h_t^S, \emptyset;\ \theta),$ & $f(p_t, h_t^S, h_{t-1}'';\ \theta)\}$ &  $\pi_t'' = g(h_t'';\ \phi'')$ \\

 & & \underline{if $t\bmod{\tau}=1$}: & \underline{otherwise}: & \\
 Perception & $h_t'=$ & $\{f(x_t, \emptyset, \emptyset;\ \theta),$ & $f(x_t, \emptyset, h_{t-1}';\ \theta)\}$ &  $\pi_t' = g(h_t';\ \phi')$ \\
\bottomrule
\end{tabular}
\end{center}
\end{table}

\paragraph{Implementation.}
Although the contents of partial information $p_t$ remain flexible in our definition, deliberate selection of this quantity may be required to enable useful regularization.  For visual environments, the recurrent A3C Agent \citep{mnih2016asynchronous} suggests a convenient delineation, which we use in our experiments: the partial observation consists of the previous action and reward in the environment, $p_t=(a_{t-1}, r_{t-1})$. This is compared to the full observation provided to the agent $x_t=(o_t, a_{t-1}, r_{t-1})$ which additionally includes the screen pixels $o_t$.  The actions provide critical information for the forward model of the \emph{prediction} branch, and they hold natural appeal as information internal to the agent. 

In practice, the PPR architecture is implemented as a self-contained recurrent neural network core, and training only requires an additional loss term $\mathcal{L}_{aux}$ computed on the current training batch, allowing the agent to be easily incorporated in most existing deep RL frameworks.  In our experiments we found it possible to use the same recurrent network weights in all branches, as reflected in Table~\ref{tab:recur_equations}.  Weight sharing limits the increase in recurrent parameters to only $2\times$ over the flat agent, using the same core size.\footnote{In our experiments, we found no effect in flat agents from using $2\times$ LSTM parameters.}

\section{Related Work}

Recurrent networks with multiple time scales have appeared in numerous forms for supervised learning of long sequences, with the intent of leveraging the prior that time-series data is hierarchical. In the Sequence Chunker ~\citep{schmidhuber1992learning}, a high recurrent level receives a reduced description of the input history from a low level, and the high level operates only at time-steps when the low level is unable to adequately predict its inputs.  Hierarchical recurrent networks ~\citep{el1996hierarchical} were constructed with various fixed structures for multiple time scales.  Hierarchical Multiscale RNNs ~\citep{chung2016hierarchical} extended this idea to include learnable hierarchy by allowing layers in a stacked RNN to influence temporal behavior of higher layers.  They introduce three possible operations in their modified LSTM: FLUSH--feed output to higher level and reset recurrent state, UPDATE--receive input from lower level and advance recurrent state, and COPY--propagate the exact recurrent state.  Our architecture can be understood in these terms, but we utilize a specific, new structure with multiple low levels assuming different roles, implemented by a fixed choice of when to perform each operation.  Clockwork RNNs (CW-RNNs)~\citep{clockworkRNN} and Phased LSTMs~\citep{phasedLSTM} perform hierarchical learning by using a range of fixed timescales for groupings of neurons within a recurrent layer. In contrast to these two methods, we construct a distinct routing of information between components (\textit{e.g.}, in CW-RNNs, information flows generically from slow to fast groups), and we require only two time scales to be effective. 

Relative to vanilla RNNs, several works have improved learning on long sequences by introducing new recurrence formulas to address vanishing gradients or provide use of explicit memory.  Long Short-Term Memory (LSTM) ~\citep{hochreiter1997long} is a widely-used standard which we employ in all our experiments.  More recent developments include: the Differentiable Neural Computer~\citep{graves2016hybrid}, an explicit memory-augmented architecture; Relational RNNs~\citep{santoro_relational}, which supplement LSTMs to include memory-interaction operations; plastic neural networks using Hebbian learning ~\citep{plasticity}; and the Gated Transformer-XL (GTrXL)~\citep{parisotto2019stabilizing}, which adapted a purely self-attention based approach~\citep{vaswani2017attention} for RL.  The GTrXL agent was measured on a similar benchmark to ours (although using a more recent learning algorithm) and showed similar or better improvements over a 3-layer LSTM agent.  However, this improvement came at the cost of orders of magnitude more parameters in their 12-layer self-attention architecture, and the other architectures likewise share the drawback of significantly increased computational burden over LSTMs.  Still, any of them could be employed as the recurrent core within the PPR agent---future work could seek compounding gains.

In RL specifically, our work relates closely to the For-The-Win (FTW) agent of~\citep{jaderberg2018human}.  The FTW agent features a slow-ticking and a fast-ticking core, similar to what is depicted in Figure~\ref{fig:recurrent}~(b), and includes a prior to regularise the hidden state distribution between slow and fast cores via an auxiliary loss.  Our work also builds on recent approaches to learning priors with information asymmetry for RL. In~\citep{galashov2018information}, a ``default'' policy is learned simultaneously to the agent's policy through a regularisation loss based on their KL divergence, which replaces entropy regularisation against the uniform prior.  In their experiments, hiding certain input information from the default policy, such as past or current states, was beneficial to the agent's learning.  Distral~\citep{teh2017distral} promotes multitask learning by distillation and regularisation of each task policy against a shared, central policy.  

Other works utilise a combination of memory modules and new learning algorithms for better learning through time~\citep{hung2018optimizing, MERLIN}, and a wealth of previous work exists on more explicit hierarchical RL which often exploits temporal priors~\citep{SuttonOptions,HeessModulated,VezhnevetsFun}.  Unlike these methods, we impose minimal change on the RL algorithm, requiring only auxiliary losses computed using components of the same form as those already present in the standard deep RL agent.


\section{Experiments}

\begin{figure}[t]
\centering
\includegraphics[width=0.9\textwidth]{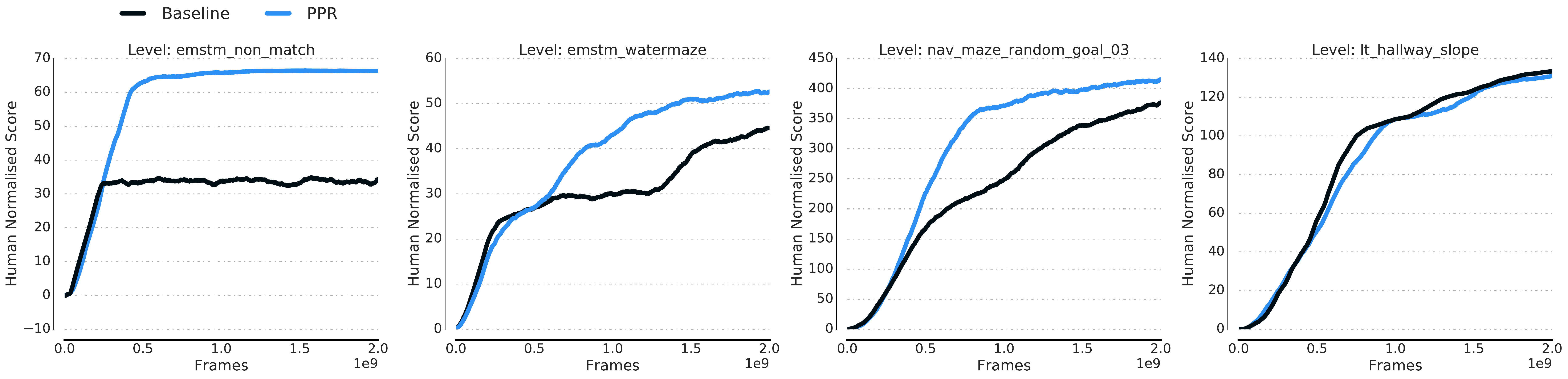}
\caption{\small Learning curves of the PPR agent (blue) compared to the baseline recurrent agent (black)~\citep{IMPALA} on four representative DMLab tasks. The PPR agent can achieve higher scores and faster learning on long-term memory tasks (\emph{e.g.} $\texttt{emstm\_non\_match}$, $\texttt{emstm\_watermaze}$, $\texttt{nav\_maze\_random\_goal\_03)}$, while not degrading in performance on more reactive tasks, such as lasertag~($\texttt{lt\_hallway\_slope}$). More levels can be found in the Appendix.}
\label{fig:fig2}
\end{figure}

We conducted experiments seeking to answer the questions: (a) does the PPR hierarchy lead to improved learning relative to flat architectures, and if so, (b) which kind of tasks is it most effective at accelerating, and (c) what are the effects of different components of the architecture. We report here experiments on levels within the DMLab-30 suite \citep{DMLab}.  It includes a collection of visually rich, 3D environments for a point-body agent with a discrete action space.  The range of tasks vary in character from memory-, navigation-, and reactive agility-based ones.   Language-based tasks are also included. Next, we report the PPR agent's performance on the recent Capture the Flag environment, which combines elements of memory, navigation, reflex, and teamwork \citep{jaderberg2018human}.  Lastly, we present  an in-depth study isolating the \emph{prediction} component of the agent in order to further clarify the effect of the combined architecture.

In our experiments, we used LSTM recurrent cores with hidden size 256 and shared weights among the three fast branches.  We trained our agents and baseline using the V-Trace algorithm ~\citep{IMPALA} on trajectory segments of length 100 agent time-steps, using action-repeat 4 in the environment.  We introduced hyperparemeters to weight each auxiliary loss, one for each branch-pair, and included these in the set of hyperparameters tuned by Population-based training (PBT)~\citep{PBT}.  For visual levels, our convolution network was a 15-layer residual network
as in ~\cite{IMPALA}, and our baselines all used the identical architecture except with a flat LSTM core for memory.  We typically fixed the slow core interval, $\tau$, to 16.

\subsection{PPR Agents in DMLab}

\paragraph{DMLab Individual Levels.}

We tested PPR agents on 12 DMLab levels. For each level, we trained a PPR and a baseline agent for 2 billion environment frames.
Figure \ref{fig:fig2} highlights results from four tasks. Compared to the baseline, the PPR agent showed significantly faster and higher learning in tasks requiring long-term memory.  In all 12 levels we tested, the PPR agent achieved the same or higher score as the baseline---full results are in the Appendix.

In \texttt{emstm\_non\_match}, the agent sees an object and must memorize it to later choose to collect any different object.  The PPR agent demonstrated proper memorisation, scoring 65 average reward, whereas the baseline agent did not, scoring 35.  In \texttt{emstm\_watermaze}, the agent is rewarded for reaching an invisible platform in an empty room and can repeatedly visit it from random respawn locations within an episode.  The second rise in learning corresponds to memorisation of the platform location and efficient navigation of return visits, which the PPR agent begins to do at roughly half the number of samples as the baseline.  The level \texttt{nav\_maze\_random\_goal\_03} is similar in terms of resets but takes place in a walled maze environment with a visible goal object.  Here as well, the PPR agent exhibits significantly accelerated learning, surpassing the final score of the baseline using less than half as many samples.  

In contrast, \texttt{lt\_hallway\_slope} is a laser tag level requiring quick reactions without reliance on long-term memory; the PPR agent is not expected to improve learning.  Significantly, the equal scores shows that the hierarchy did not degrade reactive performance.  While experimenting with agent architectures, it was a design challenge to increase performance on memory tasks without decreasing performance on reactive ones, the main difficulty being a learning mode in which the policy became less dynamic, to be easier to predict.  One effective way we found to mitigate this phenomenon is to apply $\mathcal{L}_{aux}$ to only a (random) subset of training batches (found concurrently by~\cite{chauffeurnet}), to permit the policy to sometimes update toward pure reward-seeking behavior.  Through experimentation, we also found rescaling $\mathcal{L}_{aux}$ by a factor randomly sampled from $U(0,1)$ for each batch worked well. We used these techniques in all our experiments.



\begin{figure}[t]
\centering
\begin{tabular}{cccc}
\includegraphics[width=0.26\textwidth]{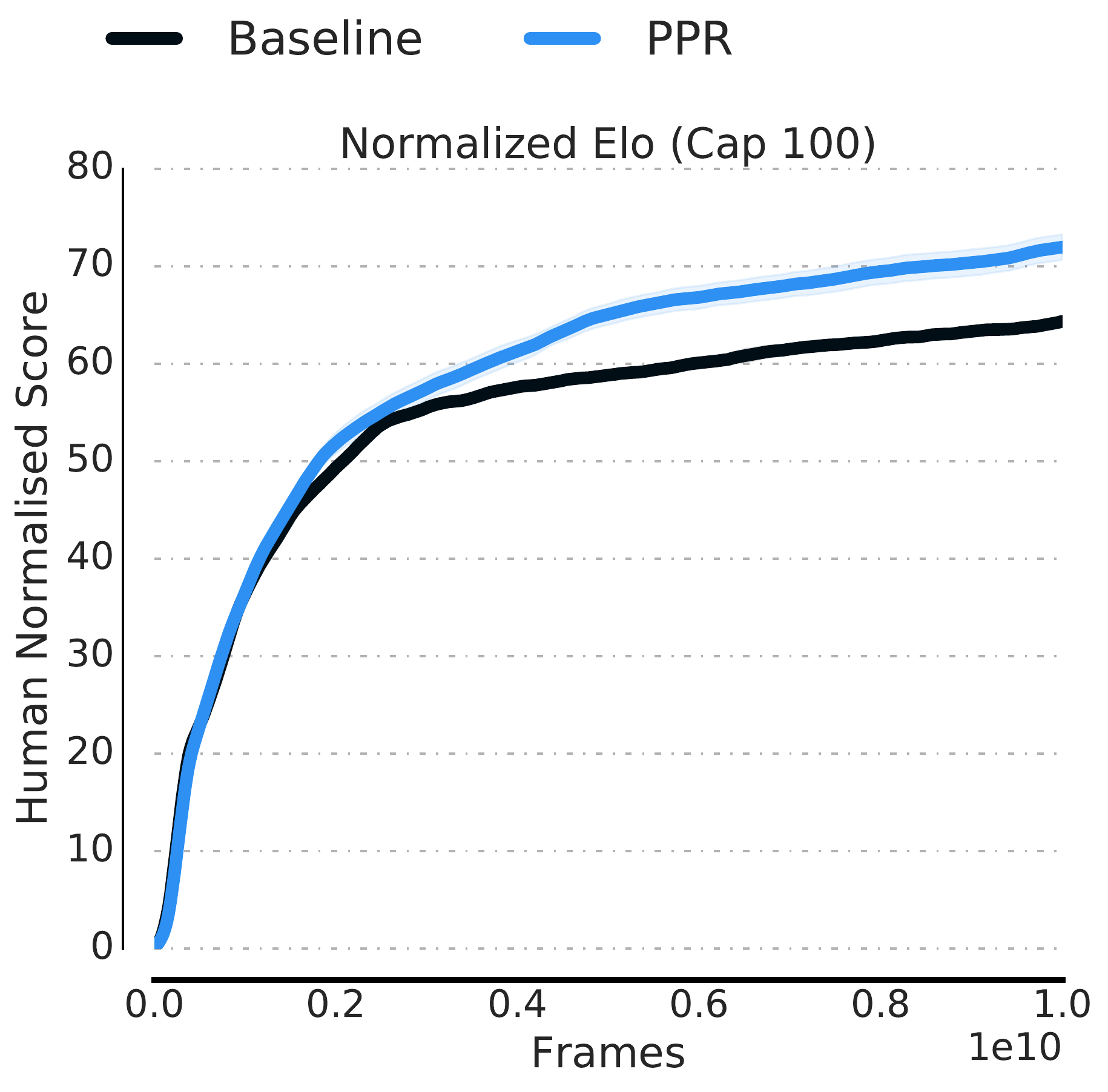} &
\includegraphics[width=0.37\textwidth]{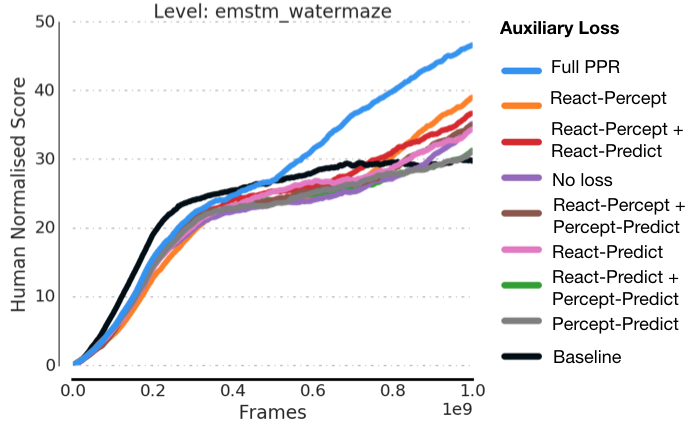} &
\includegraphics[width=0.34\textwidth]{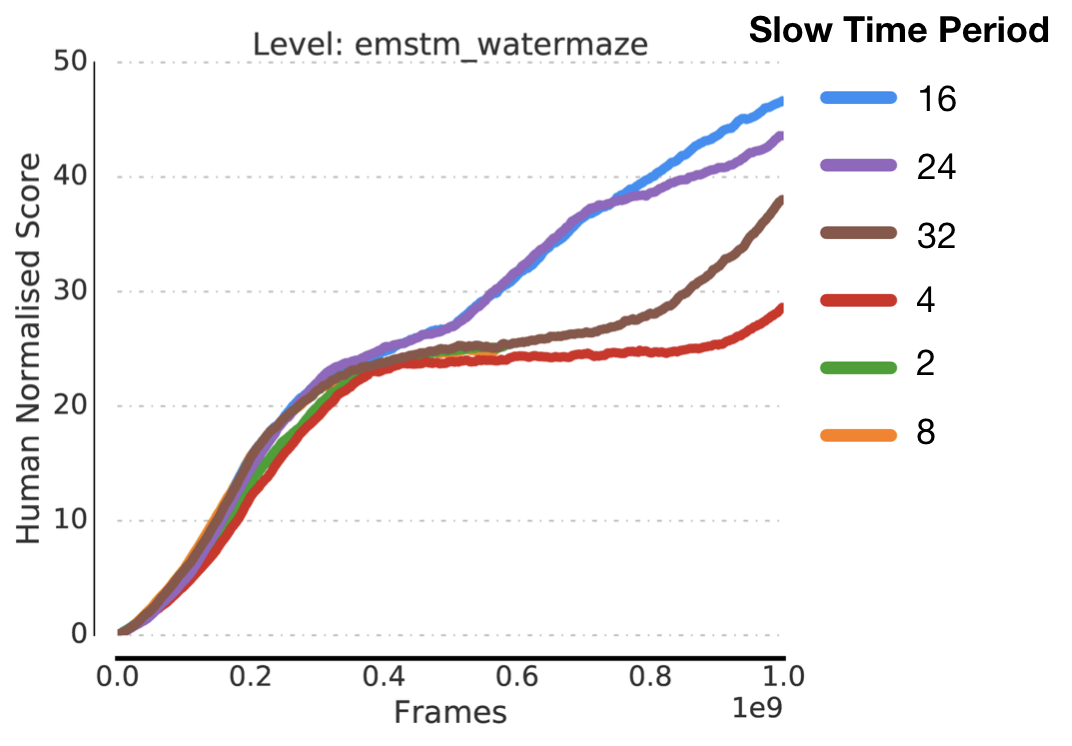}\\
(a) & (b) & (c)
\end{tabular}
\caption{\small (a) Learning curves on the DMLab-30 task domain with the PPR agent (blue) and recurrent agent baseline (black)~\citep{IMPALA}. The PPR agent consistently outperforms the Impala \citep{IMPALA} on this challenging domain. (b) Ablation study on losses. (c) Ablation study on time periods.}
\label{fig:fig3}
\end{figure}

\paragraph{DMLab-30.}
We next tested the PPR agent on a multi-task learning problem---the entire DMLab-30 suite---to test whether benefits could extend across the range of tasks while using a single set of agent weights and hyperparameters for all levels.  Indeed, the PPR agent outperformed the flat LSTM baseline, achieving an average capped human-normalized ELO across levels of 72.0\% mean (across 8 independent runs), compared to 64.3\% with the baseline~\citep{IMPALA}, Figure \ref{fig:fig3}.  The Appendix contains per-level scores from these learning runs. This difference, while modest, is difficult to achieve compared to the highly tuned baseline agent and represents a significant improvement.

\paragraph{Ablations.} To determine the effects of individual components of the PPR agent, we returned to experimenting on individual DMLab levels. First,  Figure \ref{fig:fig3}~(c) shows results from \texttt{emstm\_watermaze}, for slow core interval, $\tau$, ranging from 2 to 32.  A wide range worked well, with best performance at $\tau\geq16$.  We also experimented with evolving $\tau$ using PBT but did not observe improved performance.  

In a separate experiment, with fixed $\tau$, we activated different combinations of the three PPR auxiliary loss terms.  Using no auxiliary loss reverts to the bare Perception-Reaction architecture.  Figure \ref{fig:fig3}~(b) shows results, also on \texttt{emstm\_watermaze}, with the full PPR agent performing best.  The \emph{prediction} branch, which is only trained via the auxiliary loss, is revealed to be crucial to the learning gains, and so is inclusion of the behavior policy in $\mathcal{L}_{aux}$.  Although using two of the three auxiliary losses was sometimes effective in our experiments, we measured more consistent results with all three active.



\subsection{PPR Agents in Capture the Flag}  
\begin{wrapfigure}{R}{0.42\textwidth}
  \begin{center}
    \includegraphics[trim=10 10 10 10,clip,width=0.42\textwidth]{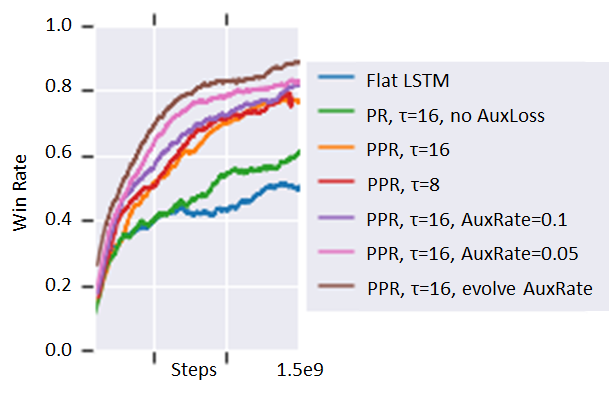}
  \end{center}
  \caption{\small Win rates against bots (skill level 4) on Capture the Flag procedural levels, of baseline recurrent agent (blue), PR agent without auxiliary losses (green), and various settings of PPR agents. }
  \label{fig:ctf_botskill4}
\end{wrapfigure}

The Capture the Flag environment is a first-person, 2-vs-2 multiplayer game based on the Quake III engine, developed in~\cite{jaderberg2018human}.  The RL agent controls an individual player, and must learn to coordinate with a teammate to retrieve a flag from the opponent base, while the opponent team attempts to do the same.  Players can ``tag'' opponents (as in laser tag), removing them from the game temporarily until they respawn at their base.  Human-level performance by RL was first achieved in this game by~\cite{jaderberg2018human}.  They trained agents from scratch using a combination of techniques including PBT, careful opponent selection for playing thousands of matches in parallel, and a temporally hierarchical agent architecture with an associated auxiliary loss and Differentiable Neural Computer recurrent cores \citep{graves2016hybrid}.  The PBT-evolved parameters included internal agent weightings for several possible reward events, to provide denser reward than only capturing a flag or winning/losing a match, which lasts 5 minutes.  Together, these advancements comprised their For-The-Win (FTW) agent. 

In our experiment, we used all components as in the FTW agent and training, except for the recurrent structure.  Our baseline used the flat LSTM architecture, and the PPR agent used LSTM cores.  Figure~\ref{fig:ctf_botskill4} shows win rates against bots, evaluated during training on the procedurally generated levels (see Appendix for bots of other skill levels).  Without auxiliary losses, the Perception-Reaction hierarchical agent performed slightly above the baseline, reaching 62\% and 50\% win rates, respectively.  By including the auxiliary losses, however, the full PPR agent dramatically accelerated learning and reached higher asymptotic performance, nearly 90\% by 1.5B steps.  Slow time scales of either 8 or 16 gave similar performance, showing low sensitivity to this hyperparameter.  Using a reduced auxiliary loss rate (0.1 and 0.05 shown) further improved performance, with the best learning resulting from evolving the rate by PBT.  In this experiment, the PPR agent demonstrated improved learning of the complicated mix of memory, navigation, and precision-control skills required to master this domain.

\subsection{Flat, Prediction Agent (Ablation)}

\begin{wrapfigure}{L}{0.35\textwidth}
  \begin{center}
    \includegraphics[width=0.33\textwidth]{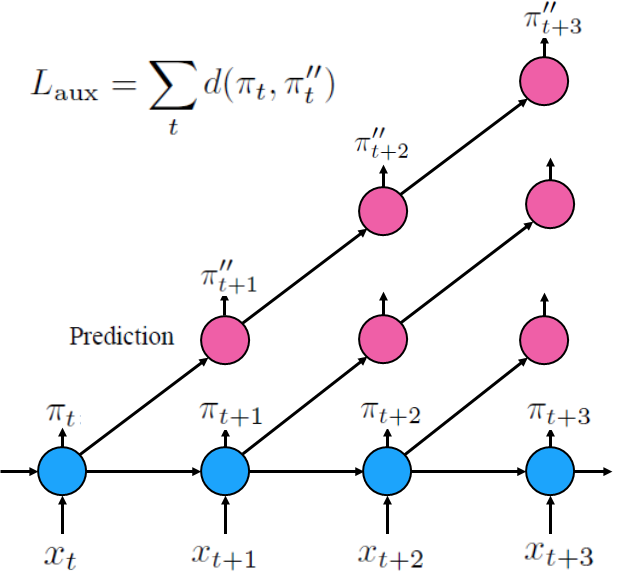}
  \end{center}
  \caption{\small Prediction agent architecture and training loss.}
  \label{fig:prediction_agent}
\end{wrapfigure}

\begin{figure}[h]
\centering
\begin{tabular}{ccc}
\includegraphics[width=0.38\textwidth]{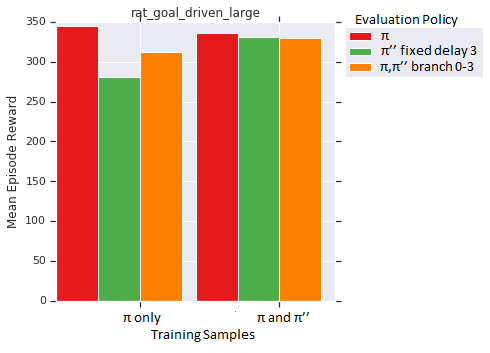} &
\includegraphics[width=0.38\textwidth]{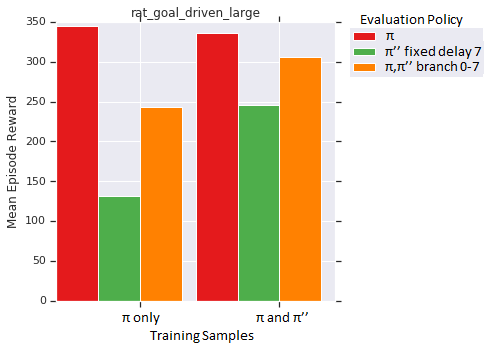} &
\includegraphics[width=0.13\textwidth]{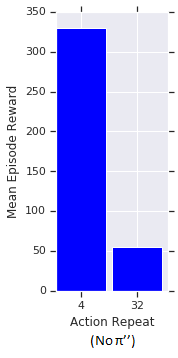}\\
(a) & (b) & (c)
\end{tabular}
\caption{\small Final evaluation scores for trained flat, \emph{prediction} agent in $\texttt{rat\_goal\_driven\_large}$. Various schemes used for drawing the behavior policy from the \emph{prediction} auxiliary policy, many of which perform similarly to the baseline (reactive) agent. (a) Agents executing \emph{prediction} policy up to 3 time steps without new observations.  (b) The same trained agents, but evaluated up to 7 time steps without new observations.  (c) Long action-repeat trained agent.}
\label{fig:flat_predictive}
\end{figure}

In another ablation, we sought performance gains for a flat LSTM agent by training with the auxiliary regularisation loss of a \emph{prediction} branch, Figure~\ref{fig:prediction_agent}.  During training we rolled out predictions up to 10 steps, and for some agents we included training samples from the prediction policy, $\pi''$.  We then evaluated final agent performance under three different behavior schemes: i) the baseline using only $\pi$, ii) using $\pi''$ at a fixed number of steps after branching, and iii) using $\pi''$ along a branch from its starting point.  Figure~\ref{fig:flat_predictive}~(a) and~(b) show evaluations using 3-step and 7-step predictions, respectively, using the same trained agents, in \texttt{rat\_goal\_driven\_large}.\footnote{This environment is similar to \texttt{nav\_maze\_random\_goal\_03} from above.}  Using the branch-following scheme, the best 7-step \emph{prediction} agents scored above 300, close to the baseline agent (around 340).  In contrast, additional baseline agents we trained with frame-skip 32 performed significantly worse---score 50, Figure~\ref{fig:flat_predictive}~(c)---despite using the same refresh rate for incorporating new observations into the policy.

These experiments show that the fast perception and control loops are essential, although they can operate more loosely coupled than in the baseline agent.  Clearly, the auxiliary policy encodes future sequences of high-reward behavior, despite lacking access to input observations over similar time scales as used for the PPR agent's hierarchy.  Yet in no case did we observe any improvement in the base agent's learning.  Evidently, the full PPR agent is needed to accelerate learning.

\section{Conclusion}

In this paper we introduced a new agent to learn in partially observable environments, the PPR agent, which incorporates a temporally hierarchical recurrent structure, as well as imposing priors on the behaviour policy to be both predictable from long-term memory only, and from current observations only. This agent was evaluated on a diverse set of 3D partially observable RL problems, and showed improved performance, in particular on tasks involving long-term memory. We ablated the various components of the agent, demonstrating the efficacy of each. We hope future work can build upon these ideas and continue exploring structural- and loss-based priors to further improve deep RL in partially observable environments.

\section*{Broader Impact}

The immediate societal impacts of this work are limited, to the extent that the methods herein remain bound to RL tasks tailored for research and in the virtual setting.  Further in the future, success of ours and related methods at producing agents capable of reasoning over extended time periods could influence how decisions are made in a variety of possible systems, hopefully with the effect of improving efficiency.  We advocate no particular applications in this research.  

A more immediate impact is the environmental cost of running our experiments, primarily the atmospheric emissions of greenhouse gases due to electricity usage.  Our experiments were of moderate but non-negligible scale, as our methods apply to sophisticated RL tasks utilizing CPU-based simulators and neural networks running on more energy-efficient GPU/TPU hardware.  Our research aims to reduce sample complexity in RL, permitting fewer compute cycles to be used for future developments.  Otherwise, we rely on broader efforts to mitigate effects at the data-center level, such as the use of non-emitting, renewable energy sources.  Perhaps the greatest risk is that future RL techniques are capable enough to engender widespread adoption, but without significantly improved computational efficiency, resulting in a net increase in overall energy usage and emissions.

\begin{ack}
Adam Stooke gratefully acknowledges previous support from the Fannie and John Hertz Foundation, and the authors thank Simon Osindero, Sasha Vezhnevets, Nicholas Heess, Greg Wayne, and others for insightful research discussions.  Compute resources were provided by DeepMind.

\end{ack}

\bibliography{main}
\bibliographystyle{main}

\newpage
\appendix
\section*{APPENDIX: Additional Learning Curves}

\begin{figure}[h]
\centering
\includegraphics[width=1.0\textwidth]{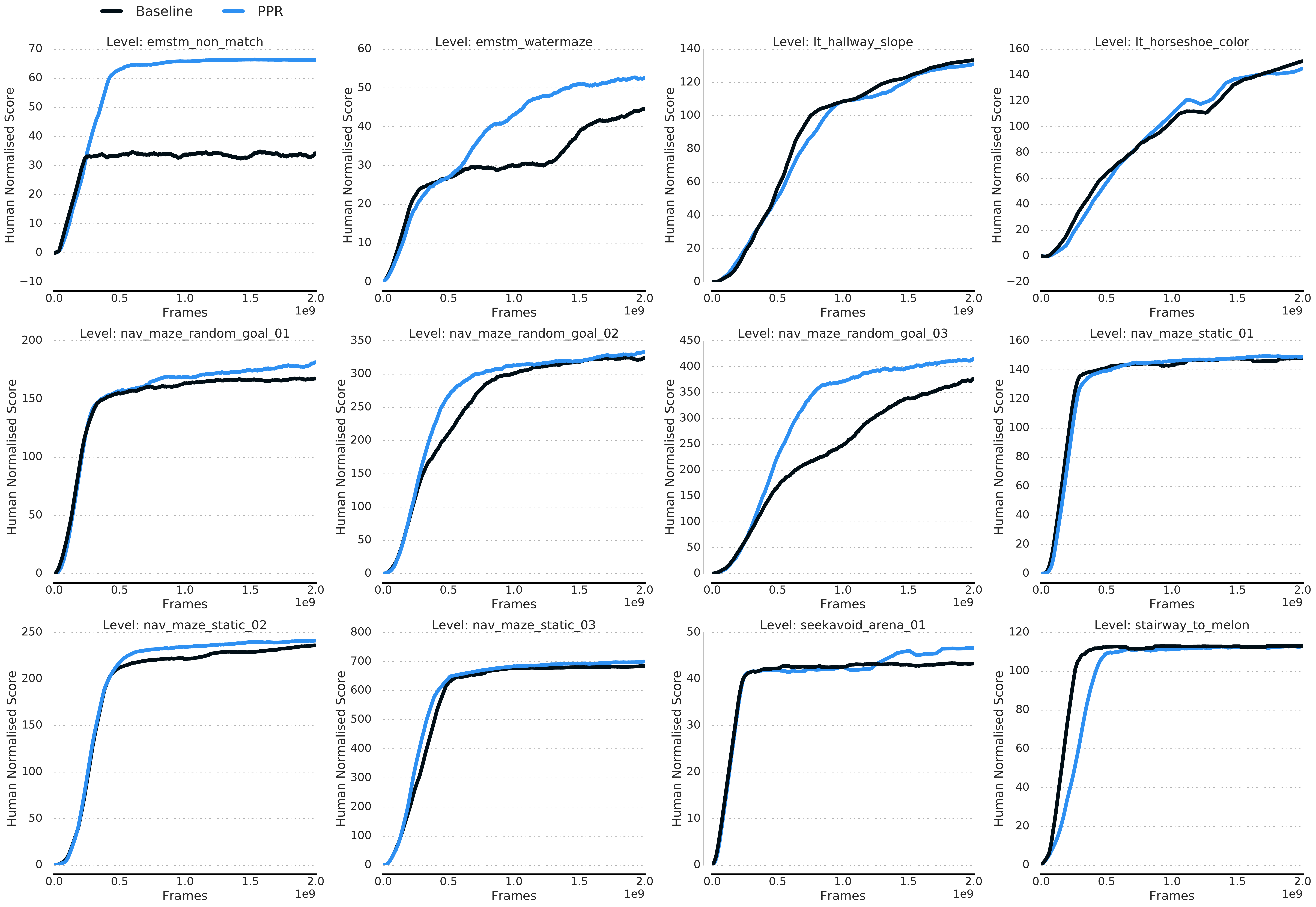}
\caption{\small Learning curves of the PPR agent (blue) compared to the baseline recurrent agent (black)~\citep{IMPALA} on various individual DMLab tasks.}
\label{fig:fig2extended}
\end{figure}

\begin{figure}[p]
\centering
\includegraphics[width=0.85\textwidth]{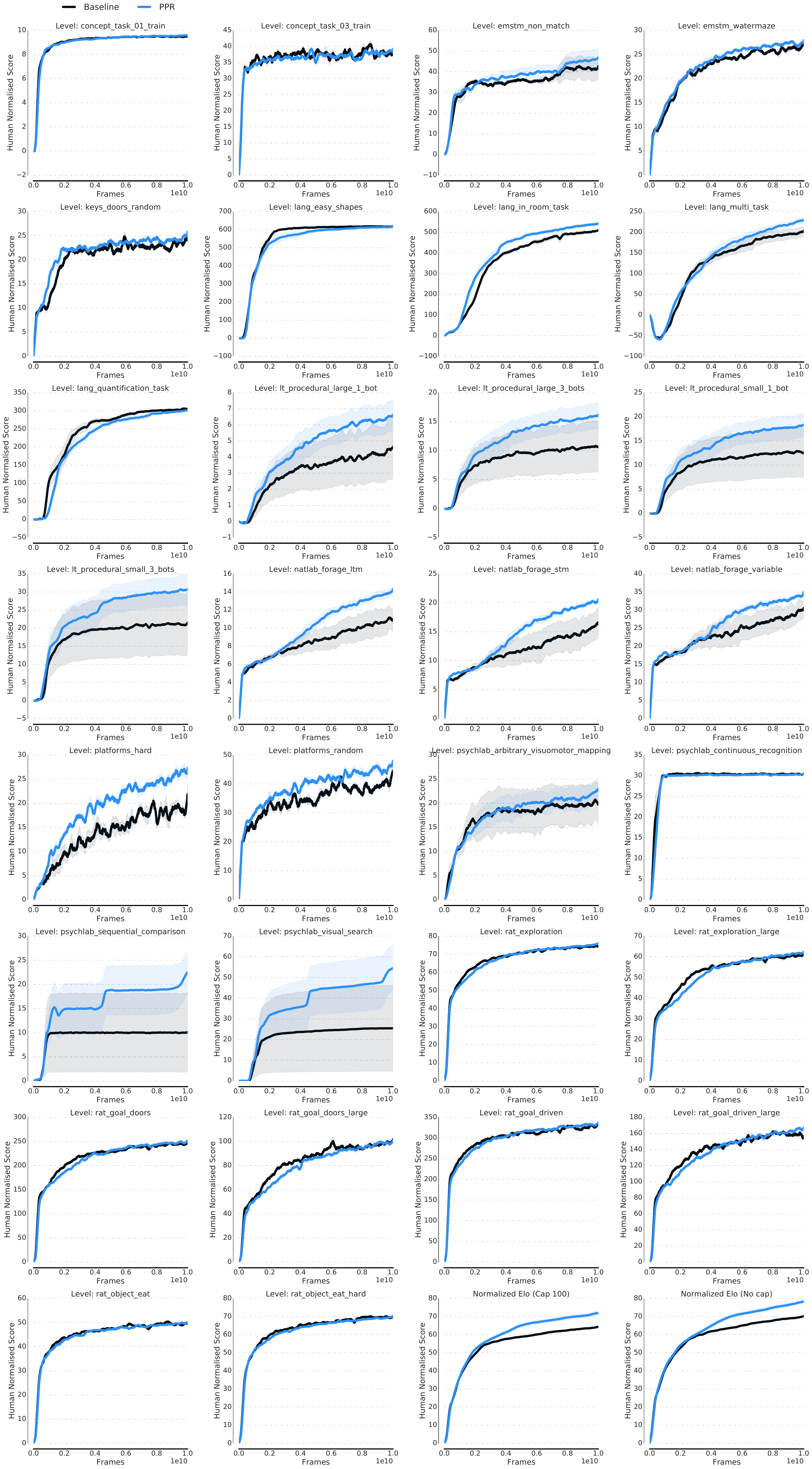} 
\caption{\small Learning curves on the DMLab-30 task domain with the PPR agent (blue) and recurrent agent baseline (black), separated by level. Shaded area shows the mean standard error. The PPR agent consistently outperforms the baseline~\cite{IMPALA} on this challenging domain.}

\label{fig:DMLab_All}
\end{figure}

\begin{figure}[p]
\centering
\includegraphics[width=1.\textwidth]{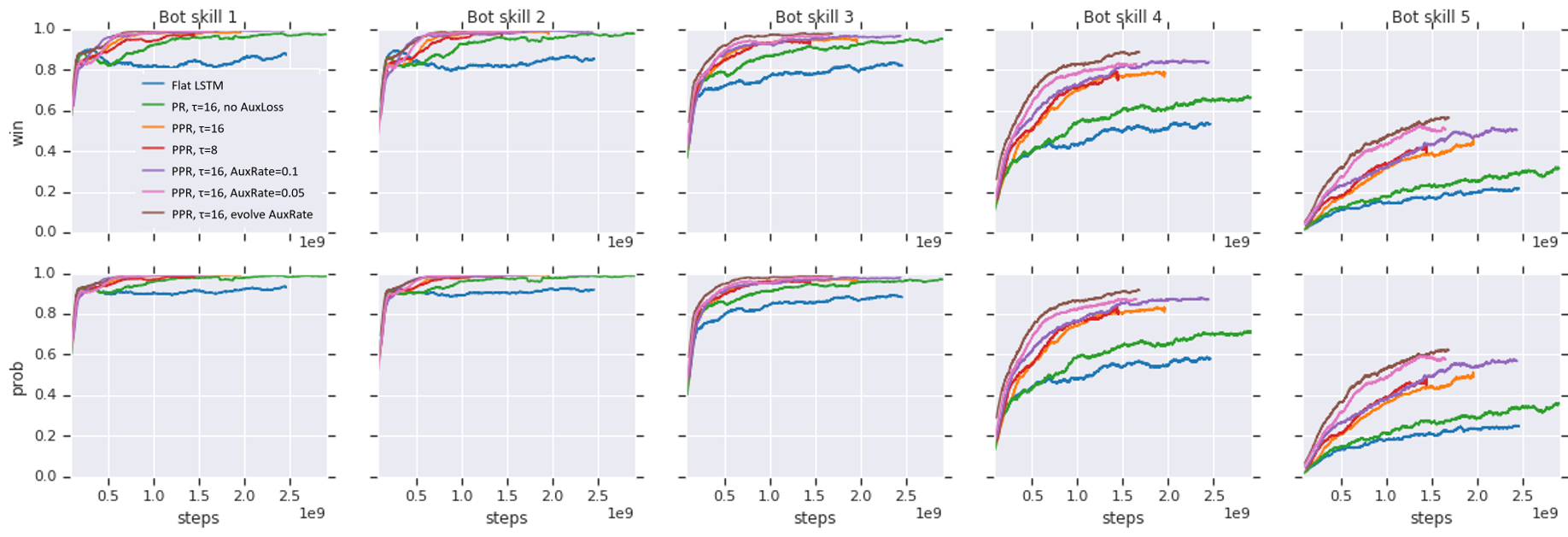} 
\caption{\small Learning curves on Capture the Flag procedural levels, evaluated against bot skill levels of increasing difficulty.  Various settings of the PPR agent outperform the PR agent (without auxiliary losses; green) and the baseline flat agent (green) in this multi-faceted domain.  Top row: win rate, bottom row: win plus one-half draw rate.}

\label{fig:ctf_all}
\end{figure}

\end{document}